\documentclass[runningheads]{llncs}

\usepackage{eccv}

\usepackage{eccvabbrv}

\usepackage{graphicx}
\usepackage{booktabs}

\usepackage[accsupp]{axessibility}  % Improves PDF readability for those with disabilities.

\usepackage{hyperref}

\usepackage{orcidlink}
\usepackage{bm}
\usepackage{url}
\usepackage{pifont}
\usepackage{bbding}
\usepackage{comment}
\usepackage{caption}
\usepackage{amsmath}
\usepackage{pdfpages}
\usepackage{multirow}
\usepackage{makecell}
\usepackage{amsfonts}
\usepackage[misc]{ifsym}
\usepackage{pifont}
\usepackage{xcolor,colortbl}
\usepackage{graphicx}
\usepackage[normalem]{ulem}
\usepackage{float} % 引入float包
\useunder{\uline}{\ul}{}

\begin{document}

% ---------------------------------------------------------------
% TODO REVIEW: Replace with your title
\title{UniMD: Towards Unifying Moment Retrieval and Temporal Action Detection} 

% TODO REVIEW: If the paper title is too long for the running head, you can set
% an abbreviated paper title here. If not, comment out.
\titlerunning{UniMD: Unified Moment Detection}

% TODO FINAL: Replace with your author list. 
% Include the authors' OCRID for the camera-ready version, if at all possible.
% \author{Yingsen Zeng\inst{1}\orcidlink{0000-1111-2222-3333} \and
% Yujie Zhong\thanks{Corresponding author}\inst{1}\orcidlink{1111-2222-3333-4444} \and
% Chengjian Feng\inst{1}\orcidlink{2222--3333-4444-5555} \and 
% Lin Ma\inst{1}\orcidlink{2222--3333-4444-5555}}

\author{Yingsen Zeng \and
Yujie Zhong\thanks{Corresponding author} \and
Chengjian Feng \and 
Lin Ma}

% \author{Yingsen Zeng \and
% Yujie Zhong\textsuperscript{\Letter}
%  \and
% Chengjian Feng \and 
% Lin Ma}

% TODO FINAL: Replace with an abbreviated list of authors.
\authorrunning{Y.~Zeng et al.}
% First names are abbreviated in the running head.
% If there are more than two authors, 'et al.' is used.

% TODO FINAL: Replace with your institution list.
\institute{Meituan Inc.
% \email{yingsen\_2@163.com} \and
% ABC Institute, Rupert-Karls-University Heidelberg, Heidelberg, Germany\\
% \email{\{abc,lncs\}@uni-heidelberg.de}
}

\maketitle

% \begin{abstract}
%   The abstract should summarize the contents of the paper. 
%   LNCS guidelines indicate it should be at least 70 and at most 150 words.
%   Please include keywords as in the example below. 
%   This is required for papers in LNCS proceedings.
%   \keywords{First keyword \and Second keyword \and Third keyword}
% \end{abstract}

\begin{abstract}

Temporal Action Detection (TAD) focuses on detecting pre-defined actions, while Moment Retrieval (MR) aims to identify the events described by open-ended natural language within untrimmed videos. Despite that they focus on different events, we observe they have a significant connection. For instance, most descriptions in MR involve multiple actions from TAD. In this paper, we aim to investigate the potential synergy between TAD and MR. 
Firstly, we propose a unified architecture, termed \textbf{Uni}fied \textbf{M}oment \textbf{D}etection (\textbf{UniMD}), for both TAD and MR. It transforms the inputs of the two tasks, namely actions for TAD or events for MR, into a common embedding space, and utilizes two novel query-dependent decoders to generate a uniform output of classification score and temporal segments. 
Secondly, we explore the efficacy of two task fusion learning approaches, pre-training and co-training, in order to enhance the mutual benefits between TAD and MR. 
Extensive experiments demonstrate that the proposed task fusion learning scheme enables the two tasks to help each other and outperform the separately trained counterparts. Impressively, \emph{UniMD} achieves state-of-the-art results on three paired datasets Ego4D, Charades-STA, and ActivityNet. 
Our code is available at \href{https://github.com/yingsen1/UniMD}{https://github.com/yingsen1/UniMD}. 

\end{abstract}
\section{Introduction}
\label{sec:intro}

Temporal Action Detection (TAD) and Moment Retrieval (MR) are two similar tasks that seek to identify specific events and their corresponding temporal segments within untrimmed videos. TAD focuses on identifying temporal segments related to a single action, while MR aims to identify temporal segments that align with a natural language description. These tasks have significant connections, for instance, most natural language descriptions in MR are a combination of multiple actions from TAD. However, most existing works \cite{zhang2022actionformer,dai2021ctrn,lei2021detecting,zhang2020span,yan2023unloc,li2024detal} treat TAD and MR as two independent tasks, employing separate models for each. In this paper, we aim to answer a question: \textbf{whether the two tasks can benefit each other by fusing them using a single model?}

\begin{figure}[ht]
  \centering
  % 第一行子图
  \begin{subfigure}[b]{0.45\linewidth}
    \hspace{-2.5mm}
    \includegraphics[height=0.70\linewidth, width=1.0\linewidth]{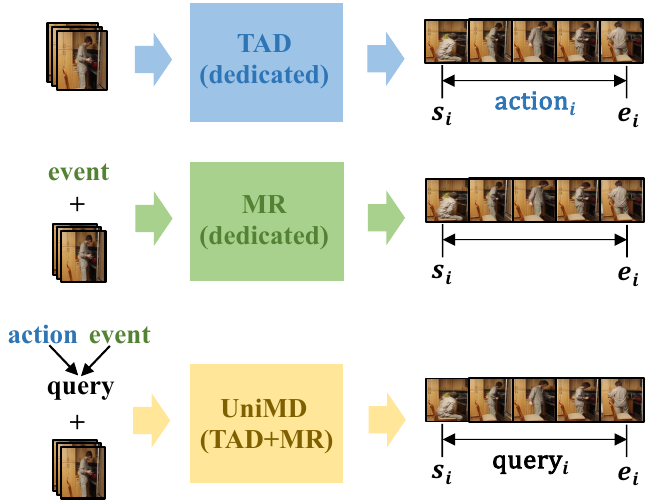}
    \caption{Illustration of TAD, MR, and UniMD. UniMD performs both TAD and MR in a single model (\ie same model parameters).}
    \label{fig:intro2_sub1}
  \end{subfigure}
  % \hfill % 在子图之间添加一些空白
  \hspace{5mm} % 增加水平间距
  \begin{subfigure}[b]{0.48\linewidth}
    % \hspace{10mm} % 手动调整
    \hfill
    \includegraphics[width=0.88\linewidth]{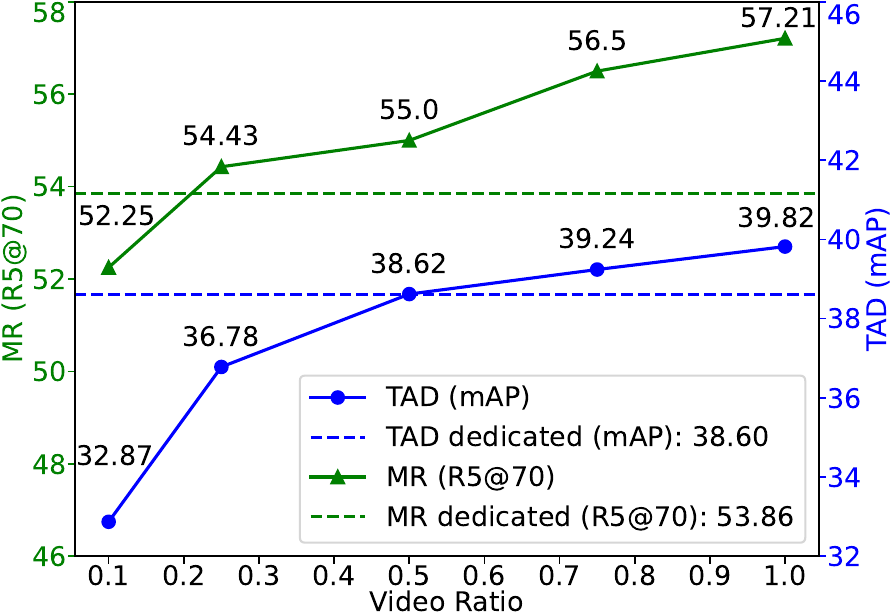}
    \hfill
    % \vspace{0.5mm}
    \caption{Co-train results evaluated on ActivityNet under various data volumes in the training set. 
    Dedicated models are trained separately for specific tasks, using overall videos as training data.
    % The dedicated models refer to models that are trained separately using overall videos for specific tasks.
    }
    \label{fig:intro2_sub2}
  \end{subfigure}
  % \vspace{-2.5mm}
  \caption{Our proposed model, UniMD, can simultaneously perform TAD and MR. When co-trained using a fraction of the training data, it can even achieve superior performance compared to dedicated models, such as 25\% training data in MR and 50\% in TAD.}
  \label{fig:intro_2}
  % \vspace{-5mm}
\end{figure}

Fusing TAD and MR is meaningful in two aspects, namely, it can not only lead to cost reduction in deployment but also holds the potential to enhance their overall performance.
By carefully exploring the videos that cover both tasks (\eg Charades \cite{sigurdsson2016hollywood} and Charades-STA \cite{gao2017tall}), we identify three potential mutual benefits between them: (i) events from MR can express the relationships and sequencing of multiple actions, thereby establishing dependencies among actions. For instance, in Fig.~\ref{fig:intro1_sub1}, 
the event ``person covered by a blanket awakens'' implies ``snuggling with a blanket'' and ``awakening'' occur at the same time. In Fig.~\ref{fig:intro1_sub2}, the event ``person sits down to look at a book'' indicates ``working on notebook'' happens after ``sitting at a table''. 
(ii) 
actions from TAD serve as a decomposition of a complete event, which provides more refined supervision to the MR task.
In Fig.~\ref{fig:intro1_sub3}, based on opposite semantics of ``putting'' and ``taking'', the action ``putting something on a table'' can be considered as a negative sample for the event ``person takes some food out of the bag''. Additionally, as shown in Fig.~\ref{fig:intro1_sub4}, all actions in TAD can be regarded as special events for MR, providing it with more positive samples.
(iii) The integration of TAD and MR enhances the amount of training instances. For example, Charades and Charades-STA have an average of 6.8 action instances and 2.4 event instances per video, respectively. Therefore, TAD can enrich MR with more than 200\% event instances, while MR can provide an additional 16 thousand action descriptions for TAD.

To investigate such potential synergies between TAD and MR, we propose a new task formulation termed Moment Detection (MD) that aims to address both TAD and MR tasks simultaneously, as illustrated in Fig.~\ref{fig:intro2_sub1}. For MD, we design a task-unified architecture, Unified network for Moment Detection (UniMD), with a task fusion learning approach, to enhance the performance of both tasks:

\begin{figure}[ht]
  \centering
  % 第一行子图
  \begin{subfigure}[b]{0.44\linewidth}
    \includegraphics[width=\linewidth]{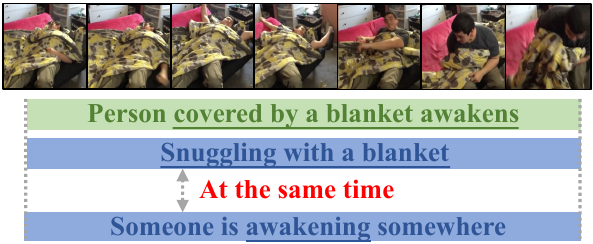}
    \caption{MR$\rightarrow$TAD: co-occurrence of actions.}
    \label{fig:intro1_sub1}
  \end{subfigure}
  \hspace{5mm} % 增加水平间距
  \begin{subfigure}[b]{0.44\linewidth}
    \includegraphics[width=\linewidth]{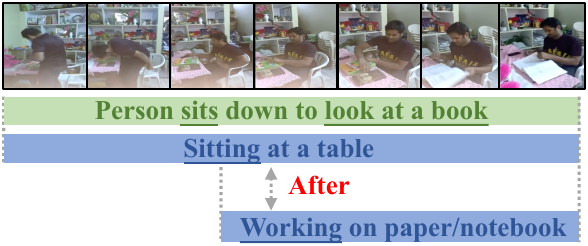}
    \caption{MR$\rightarrow$TAD: order of actions.}
    \label{fig:intro1_sub2}
  \end{subfigure}
  
  % 第二行子图
  \begin{subfigure}[b]{0.44\linewidth}
    % \vspace{3mm} % 微调子图垂直位置
    \includegraphics[width=\linewidth]{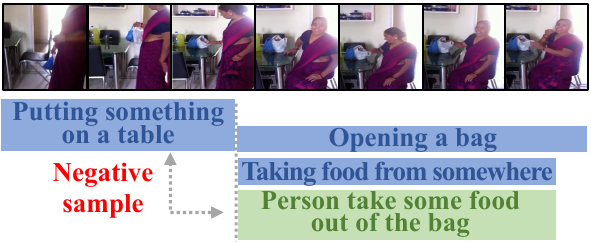}
    \caption{TAD$\rightarrow$MR: negative sample.}
    \label{fig:intro1_sub3}
  \end{subfigure}
  \hspace{5mm} % 增加水平间距
  \begin{subfigure}[b]{0.44\linewidth}
    % \vspace{-10mm} % 微调子图垂直位置
    \includegraphics[width=\linewidth]{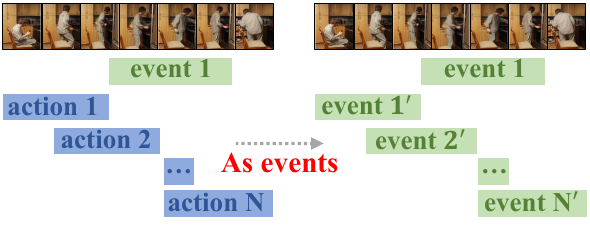}
    \caption{TAD$\rightarrow$MR: more events.}
    \label{fig:intro1_sub4}
  \end{subfigure}
  % \vspace{-2mm}
  \caption{The mutual benefits of TAD and MR tasks. The queries in green belong to \textcolor{green}{MR} and that in blue are categories of \textcolor{blue}{TAD}. The queries from MR help establish dependencies between actions like (a) co-occurrence, and (b) order. The instances from TAD can (c) act as negative samples, and (d) provide more events for MR.}
  \label{fig:intro_1}
  % \vspace{-4mm}
\end{figure}

\noindent \textbf{(1) Task-unified architecture.} Task integration poses a challenge in addressing the inconsistency in inputs (\ie only videos as input in TAD, while both videos and events as input in MR) and the disparity in action scopes (\ie TAD focuses on pre-defined actions, while MR deals with events described in natural language).
To solve this, we establish a uniform interface for task input and output. For the input, we adopt an open-ended query format such as ``a video of [action/event]'' to describe the actions and events from the two tasks. These queries are then converted into textual embeddings using a pre-trained image-text model such as CLIP \cite{radford2021learning}, \emph{which inherently establishes a relationship between actions or events}. For the output, we propose a query-dependent classification head and a query-dependent regression head to produce uniform classification scores and temporal boundaries for each query. The classification head employs textual embeddings as a classifier to generate the classification scores. Simultaneously, the regression head transforms the textual embeddings into a convolutional kernel to predict the query-relevant temporal boundaries.

\noindent\textbf{(2) Task fusion learning.} To enhance synergies between TAD and MR, we explore task fusion learning to promote their mutual influence. Specifically, we examine and discuss the impact of pre-training and co-training in task fusion learning. Moreover, we introduce two co-training methods: synchronized task sampling and alternating task sampling. 
Synchronized task sampling prioritizes video samples encompassing both tasks, ensuring each training iteration includes both tasks. In contrast, alternating task sampling updates the network based on an alternating task at each iteration. Among them, co-training with synchronized task sampling effectively enhances the synergy and leads to distinct improvements for each task. 
As shown in Fig.~\ref{fig:intro2_sub2}, the proposed co-trained model can achieve better results than dedicated models, even with only a subset of training data, \ie, 25\% training videos for MR and 50\% training videos for TAD. 
\emph{This demonstrates that the mutual benefits are not merely from the increased quantity of annotations, but rather from the enhanced effectiveness of co-training.}

To summarize, this paper makes three contributions: (i) we propose a unified framework, termed UniMD, to tackle TAD and MR tasks simultaneously. It links actions and events by inheriting the text encoder of CLIP to encode the query and predicts the action- or event-relevant temporal boundaries via the query-dependent heads. (ii) To the best of our knowledge, we are the first to exploit task fusion learning to investigate whether the two tasks can help each other and propose an effective co-training method to enhance their synergy, which provides valuable insights and experiences for training large language-based video models (\eg LLaMA-vid~\cite{li2023llama}) on action/event detection. (iii) Extensive experiments demonstrate that UniMD achieves state-of-the-art performance across various benchmarks: namely, 23.25\% mAP in Ego4D-MQ, 14.16\% R1@30 in Ego4D-NLQ, 63.98\% R1@50 in Charades-STA, and 60.29\% mAP@50 in ActivityNet.

\section{Related Work}
\label{sec:related_work}

\textbf{Temporal Action Detection (TAD).} TAD is dedicated to the identification of predefined actions and their temporal segments in untrimmed videos. It can be classified into two-stage methods \cite{zeng2019graph,zhu2021enriching,zhao2017temporal} and one-stage methods \cite{zhang2022actionformer,lin2021learning,lin2017single}. Two-stage methods initially generate video segments as action proposals, which are subsequently categorized into actions and refined in terms of their temporal boundaries. Conversely, one-stage methods aim to localize actions within a single shot without relying on action proposals. Some TAD tasks involve a substantial number of action categories densely labeled in videos. To tackle this challenge, several approaches \cite{dai2021ctrn,sardari2023pat,dai2021pdan,dai2022ms} have been developed to strengthen the dependencies between actions and capture both short-term and long-term dependencies. 

% ------------------------------------

\noindent\textbf{Moment Retrieval (MR).} Unlike TAD, MR aims to identify specific actions using open-ended natural language descriptions. MR can be categorized into one-stage and two-stage methods. Two-stage methods \cite{gao2017tall,zhang2019man,yuan2019semantic,zhang2020learning}, also known as proposal-based methods, utilize sliding windows and proposal generation models to generate a set of candidate windows. These candidates are then ranked, and the one with the highest similarity to the given description is selected as the final result. One-stage methods \cite{yuan2019find,zeng2020dense,mun2020local,liu2022umt}, also called proposal-free methods, directly compute the time pair or output confidence scores for each video snippet. Since MR searches for open-ended events rather than closed-set categories, it is crucial to fully utilize the semantic information within the queries. Therefore, several methods \cite{yuan2019find,mun2020local,chen2020rethinking,zhang2020span} 
focus on incorporating multi-modal features and enhancing cross-modal interactions to improve boundary regression.

% ------------------------------------

\noindent\textbf{Vision-Language Modals.} Studies on the interaction of vision and languages have been ongoing for several years. Recently, some methods \cite{radford2021learning,jia2021scaling,yao2021filip} scale up training processes by leveraging large-scale datasets, resulting in the progress of robust visual-text alignment. Furthermore, many efforts \cite{wang2021actionclip,wang2023internvid} have been made to extend these methods to video-text modals. Consequently, an increasing number of researches \cite{zou2023generalized,yan2023universal,kirillov2023segany} focus on effective techniques for fusing vision-language representations and utilizing them to establish connections across various tasks, aiming for a universal objective.

\begin{figure*}[t]
		\centering
		\includegraphics[width=12.0cm]{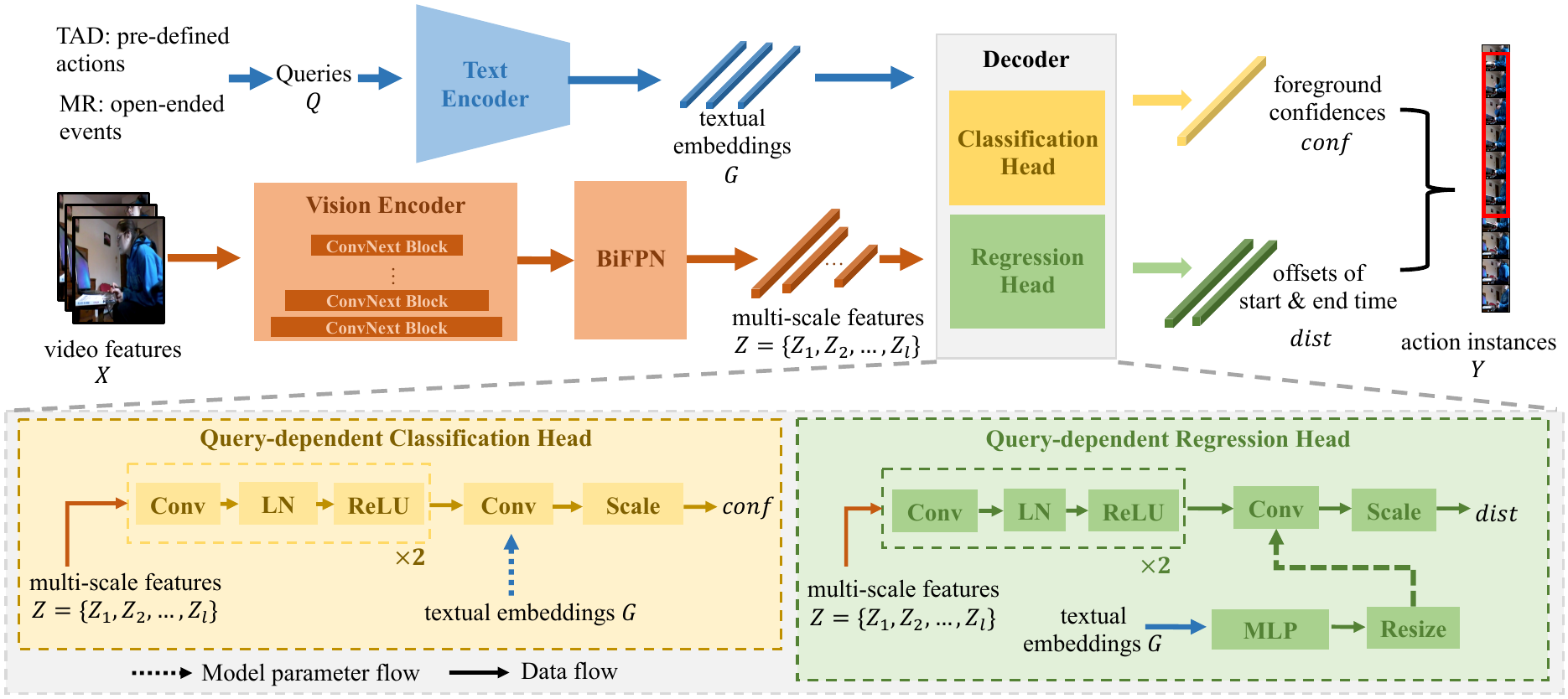}
		% \vspace{-3mm}
            \caption{\textbf{Overview of UniMD.} 
            The network is designed to process moment detection by treating each TAD category as an independent natural language query. 
            The video features are fed into the vision encoder and BiFPN to draw multi-scale features.
            The text embeddings are then streamed into the decoder, 
            enabling the calculation of foreground confidence for each time step and the onset and offset of the actions. The classification head utilizes textual embeddings as classifiers and the regression head employs the transformation of textual embeddings as convolutional kernel.
            }
		\label{demo:network}
            % \vspace{-3.0mm}
\end{figure*}

% intro-mutual_benefit
\section{Problem Definition}
\label{subsec:problem}
To investigate the potential synergies between Temporal Action Detection (TAD) and Moment Retrieval (MR), we propose a unified task formulation, called Moment Detection (MD), which addresses both sub-tasks simultaneously. In this section, we first present the problem definitions of TAD and MR, and then introduce the MD task.

TAD focuses on identifying temporal segments of a pre-defined set of \textbf{action categories} in untrimmed videos, as shown in the first row of Fig. \ref{fig:intro2_sub1}. Formally, given $C$ action categories, a TAD function $f_{\text{tad}}$ takes video images/features $X$ as input and generates the temporal segments of the action categories:
\begin{equation}
    \{(s, e, c_{\text{tad}})\}=f_{\text{tad}}(X),
\end{equation}
where $s$ and $e$ denote the start time and end time of the action category $c_{\text{tad}}\in[1, C]$. Different from TAD, as shown in the second row of Fig. \ref{fig:intro2_sub1}, MR aims to predict the temporal segments of the \textbf{events} described by the open-ended natural language:
\begin{equation}
    (s_{i}, e_{i})=f_{\text{mr}}(X, {event}_{i}),
\end{equation}
where $f_{\text{mr}}$ denotes a MR function, and $s_{i}$ and $e_{i}$ denote the start time and end time corresponding to the $i\text{-th}$ description ${event}_{i}$. 

As mentioned in Sec. \ref{sec:intro}, TAD and MR involve retrieving specific temporal segments in untrimmed videos, and have the potential to mutually enhance each other. Therefore, we propose a new task, Moment Detection (MD), to simultaneously perform both sub-tasks:
\begin{equation}
    \{(s, e, c_{\text{md}})\}=f_{\text{md}}(X, \{event\}),
\end{equation}
where $f_{\text{mr}}$ denotes an MD function, $c_{\text{md}}\in[1, C + N]$ denotes the action categories and natural language descriptions, and $N$ denotes the number of the natural language descriptions from MR. Namely, moment detection is capable of identifying the temporal segments of \textbf{both the close-set actions and the open-ended events}.
In this paper, \emph{our goal is to design a unified network to perform moment detection} (Sec.~\ref{subsec:architecture}), \emph{and propose a task fusion learning approach to enhance both TAD and MR tasks} (Sec.~\ref{subsec:learning}).

\section{Task-unified Architecture}
\label{subsec:architecture}

In this section, we investigate the potential synergy between TAD and MR by integrating both tasks within a unified framework. To this end, we propose a unified network, called Unified network for Moment Detection (UniMD). It can effectively coordinate these two tasks and facilitate their mutually beneficial advancement. 
As illustrated in Fig. \ref{demo:network}, we first unify the input interface for TAD and MR. Specifically, we formulate a uniform query $q$ to describe the action category and the open-ended events that need to be identified for each video. Regarding TAD, we utilize the action names or their variations (\eg ``a video of [action name]'') as the input queries. For MR, we directly employ their natural language descriptions as the input queries. 
Additionally, following the previous works \cite{zhang2022actionformer,lei2021detecting,zhang2020span}, we divide the entire video into multiple snippets based on predefined rules, and pre-extract a sequence of video features $X=\left \{ x^1, x^2, \dots,x^T \right \}$ using the 3D video classification models \cite{carreira2017quo,feichtenhofer2020x3d,feichtenhofer2019slowfast}, where $T$ represents the number of time steps.
Subsequently, the unified model $\Phi_{\text{md}}$ takes both input queries $Q$ and video features $X$ as inputs, and outputs the uniform predictions:
\begin{equation}
    \{(s, e, conf)\}=\Phi_{\text{md}}(X, Q),
\end{equation}
where $Q=\left \{q_i | i\in[1, C+N] \right \}$, and $conf\in\mathbb{R}^{C+N}$ denotes the foreground confidence for the close-set action categories from TAD or the events from MR.

As shown in~Fig. \ref{demo:network}, the proposed unified model is inspired by object detection \cite{lin2017focal,tian2019fcos} and contains two encoders: a vision encoder $\Phi_{\text{v-enc}}$ and a text encoder $\Phi_{\text{t-enc}}$, as well as two decoders: a query-dependent classification head $\Phi_{\text{cls}}$ and a query-dependent regression head $\Phi_{\text{reg}}$, where $\Phi_{\text{v-enc}}$ denotes a combination of the vision encoder and Feature Pyramid Network (FPN). 
The two encoders encode the textual embedding and visual embedding, while the decoders predict the confidence and temporal segments corresponding to the actions or events, based on both textual and visual embeddings. 
The unified model is designed as an anchor-free and one-stage temporal detector. \emph{Its design novelty mainly lies in the two query-dependent decoders}, which will be explained in Sec.~\ref{sub:decoder}.

\subsection{Encoder}

\noindent\textbf{Text Encoder.} To establish a correspondence between the action names from TAD and the natural language descriptions from MR, we utilize the text encoder from a well-trained image-text alignment model, \eg CLIP \cite{radford2021learning} in our case, to extract the textual embedding of the input query:
\begin{equation}
    g_{i}=\Phi_{\text{t-enc}}(q_{i}),
\end{equation}
where $g_{i}$ denotes the textual embedding of the input query $q_{i}$. CLIP is typically trained on a large corpus of image-caption pairs, and has exhibited tremendous success in open-vocabulary image classification and object detection \cite{feng2022promptdet,zhong2022regionclip}. Here, we integrate CLIP into video moment detection, offering two inherent advantages. \emph{Firstly}, it naturally establishes the connection between action names and natural language descriptions. \emph{Secondly}, it enables the action/event detection to be open-ended, \ie beyond the actions/descriptions in the training phase.

\vspace{1mm}
\noindent \textbf{Vision Encoder.} 
The vision encoder takes a sequence of video features $X$ as inputs and processes them through multiple blocks that include down-sampling operations, resulting in a series of multi-scale representations. Unlike previous works in video understanding that utilize Transformers \cite{liu2022end,wang2021actionclip,ju2022prompting}, our approach employs the ConvNext \cite{liu2022convnet} block as the main component of the visual encoder.
We adopt the pure convolutional architecture without self-attention, as we assert that long-term temporal information is not vital for local action/event detection. Conversely, convolutional operations can more efficiently emphasize information from neighboring frames, thus capturing motion patterns and temporal dependencies more effectively. In conjunction with FPN, the vision encoder generates the multi-scale features as: 
% $Z=\Phi_{\text{v-enc}}(X),$
\begin{equation}
    Z=\Phi_{\text{v-enc}}(X),
\end{equation}
where $Z=\left \{Z^1, Z^2, \dots, Z^L \right \}$ and $L$ is the number of feature pyramid levels.

\subsection{Decoder}
\label{sub:decoder}

The proposed decoder is specifically designed for moment classification and localization. It takes feature pyramids $Z$ and textual embeddings $G$ as input, and generates a series of moment instances $Y=\{(s, e, conf)\}$. To enable the decoder to adapt to individual queries, we propose a query-dependent classification head and a query-dependent regression head for multi-modal fusion. As illustrated in Fig.~\ref{demo:network}, the head is a simple and lightweight convolutional network. All heads across different pyramid levels share the same structure and weights.

\vspace{1mm}
\noindent\textbf{Query-dependent classification head.} 
The proposed classification head is composed of two regular convolutional layers and one query-dependent convolutional classifier, as illustrated in Fig.~\ref{demo:network}. The query-dependent classifier computes the inner product between the temporal visual features and the textual embeddings, yielding a similarity score that reflects the semantics of the given query. The similarity is finally scaled and passed through a Sigmoid function to obtain the semantic scores between 0 and 1. To elaborate, by taking the inputs of the visual features and the semantic queries, the classification head decodes them into a sequence of foreground confidence for the corresponding queries at every time step, described as follows:
\begin{equation}
\begin{aligned}
\left\{ conf^t\mid t\in [1, T]\right \} & = \Phi_{\text{cls}}\left ( G, Z \right ) \\
& =\Phi_{\text{sigmoid}}(\Phi_{\text{scale}}(G\odot \Phi_{\text{conv}}(Z))))), \\
\end{aligned}
\end{equation}
where $\Phi_{\text{sigmoid}}$, $\Phi_{\text{scale}}$ and  $\Phi_{\text{conv}}$ denote the Sigmoid function, the Scale operation, and the first two convolutional layers, respectively.

% \vspace{1mm} %todo
\noindent\textbf{Query-dependent regression head.} 
The regression head is utilized to predict the temporal offsets for the given actions/events at each time step. In typical object detection, the regression head can be designed to be class-agnostic due to the obvious and similar boundaries of each instance \cite{feng2022promptdet}. However, in video action/event detection, the temporal boundaries are closely associated with the detection category, for example, the queries ``Opening a bag'' and ``Taking food from somewhere'' correspond to two distinct and highly action-related boundaries, as shown in Fig.~\ref{fig:intro1_sub3}. To this end, we propose a query-dependent regression head to predict the boundaries that are strongly associated with the query.

As shown in Fig.~\ref{demo:network}, the proposed regression head has a similar structure to the classification head, \ie it consists of two regular convolutional layers and one query-dependent convolutional layer. However, unlike the classification, the textual embeddings cannot be used directly as the convolutional kernel to regress the timestamps, because the text encoder (\ie CLIP) is primarily trained using contrast learning for classification purposes. To address this issue, we introduce a query transformation branch which is responsible for semantic transformation, enabling the textual embeddings to be used effectively for the regression task:
\begin{equation}
w_{\text{reg}}=\Phi_{\text{resize}}(\Phi_{\text{mlp}}(G)),
\end{equation}
where $\Phi_{\text{mlp}}$ and $\Phi_{\text{resize}}$ denote the Multi-Layer Perceptron (MLP) and the Resize operation, $w_{reg}\in\mathbb{R}^{K \times K \times CH \times 2}$ is the transformed embeddings for time regression, and $K$ and $CH$ represent the kernel size and the number of input channels of the query-dependent convolutional layer. The query-dependent convolutional layer takes $w_{\text{reg}}$ as its kernel and performs query-wise convolution to predict the temporal offsets for each category. Thus, the regression head can be formulated as follows:
\begin{equation}
\begin{aligned}
 \left\{(d_{\text{s}}^{t}, d_{\text{e}}^{t}) \mid t\in [1,T])\right\} & =\Phi_{\text{reg}}(G, Z) \\
 & =\Phi_{\text{relu}}(\Phi_{\text{scale}}(\Phi_{\text{q-conv}}(\Phi_{\text{conv}}(Z)))), 
\end{aligned}
\end{equation}
where $\Phi_{\text{q-conv}}$ and $\Phi_{\text{relu}}$ indicate the query-dependent convolution layer and the ReLU function, and $(d_{\text{s}}^{t}, d_{\text{e}}^{t})$ denotes the time distance from the start point and the end point of the action/event to the current time step $t$. The start point and the end point of the action/event can be obtained by $s^t = t - d_\text{s}^{t}$ and $e^t = t + d_\text{e}^{t}$ at each time step.

Finally, the detection results $y^t_i=(s^t_i, e^t_i, conf^t_i)$ corresponding to the $i$-th query at each time step $t$ are obtained by combining the outputs from the classification head and the regression head. These results would be filtered through classification threshold and Soft-NMS \cite{bodla2017soft} to obtain the final output $\{(s, e, c_\text{md})\}$.

\subsection{Loss Function}

We employ two loss functions, namely (i) multi-way binary classification loss $\mathcal{L}_{\text{cls}}$ and (ii) distance regression loss $\mathcal{L}_{\text{reg}}$, to train the task-unified network. 
The losses for TAD and MR can be represented as follows:
\begin{equation}
\mathcal{L}_{\text{task}}=\sum_{t}^{}\sum_{i}^{}(\mathcal{L}_{\text{cls}} + \delta_{t,i} \cdot \mathcal{L}_{\text{reg}}),
\end{equation}
where $\delta_{t,i}$ is set to 1 only for time step $t$ that serves as a positive sample corresponding to the $i$-th query. Otherwise, it is set to 0. To jointly train both tasks, we use a weighted sum to adjust the balance between the two losses. The total loss is given by: $\mathcal{L}=\lambda_{\text{tad}} \cdot \mathcal{L}_{\text{tad}} + \lambda_{\text{mr}} \cdot \mathcal{L}_{\text{mr}}, $
where the balanced weights are empirically set to $\lambda_{\text{tad}}=3$ and $\lambda_{\text{mr}}=1$.

\section{Task Fusion Learning}
\label{subsec:learning}

In this section, we aim to explore the central question of this work: whether TAD and MR can benefit each other through task fusion learning? 
In Sec.~\ref{sec:intro}, we analyse the annotations of TAD and MR, and identify three potential benefits of task fusion learning: 
(i) \textbf{MR aids TAD.} 
Most descriptions in MR depict behaviors that involve multiple actions, which helps TAD in identifying co-occurring/sequential actions by establishing a link between adjacent single actions. 
(ii) \textbf{TAD aids MR.}
TAD provides a more detailed decomposition of behavior, which in turn can lead to a more precise estimation of time intervals for the MR task.
(iii) \textbf{Benefits each other.} 
Both tasks enhance the availability of annotations, \ie TAD enriches the action instances for MR, while MR offers numerous event descriptions that can be considered as specific action categories.
In the following, we describe two task fusion learning approaches to optimize the potential and improve the performance of both tasks.

\vspace{1mm}
\noindent\textbf{Pre-training. }
The common practice in task fusion learning is to adopt the idea of transfer learning, which pre-trains a model on one task and uses that knowledge to improve the performance on another task. However, this approach requires two sets of model parameters for the two tasks.

\vspace{1mm}
\noindent\textbf{Co-training. }
Another idea for task fusion learning is co-training. In this approach, both tasks are trained simultaneously (thanks to the novel design of UniMD), starting from a randomly initialized model, to accomplish both tasks within a single training process. 
Based on different purposes, we experimented with three different sampling methods:
(i) \textbf{Synchronized task sampling} focuses on the interaction of action categories from TAD and textual semantics from MR, ensuring that each iteration involves both tasks. It first gives priority to sampling the videos that cover both tasks, and then randomly samples the pairs of TAD and MR from the rest of the videos. 
(ii) \textbf{Alternating task sampling} treats TAD and MR as a singular task and applies equal sampling to both. It alternately samples the videos from one task and updates the network based solely on a single task at each training iteration.
(iii) \textbf{Random task sampling} has no preference and randomly samples videos, so that each iteration may contain either a single task or both tasks.

In Sec.~\ref{subsec:ablation}, we present experiments on pre-training and co-training to demonstrate the influence of task fusion learning, and subsequently discuss the efficacy of the proposed sampling methods.
\section{Experiments}

\subsection{Datasets and Evaluation Metrics}

We conducted experiments on three pairs of datasets, which cover both TAD and MR tasks. The detailed data analysis is presented in supplementary material.

\noindent\textbf{Ego4D.} The Ego4D Episodic Memory benchmark \cite{grauman2022ego4d} includes two tasks: Moment Query (MQ) and Natural Language Query (NLQ), which are respectively referred to as TAD and MR in this paper.
The majority of the videos ($\sim$98\%) contain only one task. 
Our models are evaluated on the validation and test split (using EvalAI's test server). The mean Average Precision (mAP) and recall rate under various temporal Intersection over Union (IoU) thresholds are common evaluation metrics for localization tasks. For TAD, we report mAP and top-1 recall rate at 50\% IoU. Regarding MR, we use top-1 and top-5 recall rates at IoU thresholds of 30\% and 50\% respectively.

\noindent\textbf{Charades and Charades-STA.} 
Charades~\cite{sigurdsson2016hollywood} is a densely labelled action detection dataset, and Charades-STA~\cite{gao2017tall} is an extension of Charades, introducing natural language descriptions as annotations. All videos from Charades-STA are present in Charades, accounting for 67\% of Charades. We use Charades for TAD and Charades-STA for MR. For TAD, we evaluate mAP metric using the official ``Charades\_v1\_localize'' evaluation, consistent with previous works \cite{dai2021ctrn,kahatapitiya2023weakly}. 
For MR, we report top-1 and top-5 recall rates at IoU thresholds of 50\% and 70\%.

\noindent\textbf{ANet and ANet-Caption.} 
ActivityNet (ANet)~\cite{caba2015activitynet} is divided into training, validation, and test set. Following previous works, we train our models on the training set and evaluate performance on the validation set. ActivityNet-Caption (ANet-Caption)~\cite{krishna2017dense} is built upon ANet, with a significant overlap of videos ($>$99\%). It is divided into train split, val\_1 split, and val\_2 split. Following \cite{yan2023unloc}, we use val\_1 for validation and val\_2 for testing. We consider ANet as a TAD task and ANet-Caption as an MR task. For TAD, we use mAP and mAP at the IoU threshold of 50\% as evaluation metrics. For MR, we use top-5 recall rates at IoU thresholds of 50\% and 70\%.

\subsection{Implementation Details}
\label{exp:imp_details}

\textbf{Network.} 
We use the text encoder from CLIP-B \cite{radford2021learning}, with frozen parameters during training. The vision encoder includes seven ConvNext blocks. Within them, two blocks serve as a stem, while the remaining five layers 2$\times$ downsample to generate the feature pyramid. The FPN incorporates three layers of BiFPN \cite{tan2020efficientdet}. In decoder, the classification head comprises two convolutional layers and one query-dependent convolutional layer. The regression head follows the same configuration, and includes a query transformation branch with three fully connected (fc) layers.
More details can be found in supplementary material.

\vspace{1mm}
\noindent\textbf{Video features.} (1) \textbf{Ego4D}: the video features are extracted by VideoMAE~\cite{tong2022videomae} fine-tuned with K700 dataset \cite{kay2017kinetics} and verb subset \cite{lin2022egocentric} of Ego4D, which is displayed as ``K700 $\rightarrow$ Verb'' in \cite{chen2022ego4d}. The features with 1024-D are based on clips of 16 frames at a frame rate of 30 and a stride of 16 frames. (2) \textbf{Charades}: we utilize an I3D model trained on Kinetics \cite{kay2017kinetics} to draw video features for Charades and Charades-STA, using snippets with a window of 16 frames and a stride of 4. 
Each snippet yields two types of 1024-D features: one based on RGB frames decoded at 24 fps, and the other based on optical flow. 
In addition, we extract CLIP features every 4 frames using CLIP-B.
(3) \textbf{ANet}: we follow the approach \cite{wang2022internvideo} for feature extraction, the backbone of which is named InternVideo and based on ViT-H \cite{dosovitskiy2020image}. Each step of the video features is extracted from a continuous 16 frames with a stride of 16 in 30 fps video.

\vspace{1mm}
\noindent\textbf{Text embeddings.} 
In TAD, we use the category name itself as input to extract text embeddings, without additional prefixes/suffixes/learnable prompts~\cite{ju2022prompting, radford2021learning,yan2023unloc, nag2022zero}. In MR, we use natural language descriptions from annotations. 
Notably, some methods in ANet employ score fusion strategy~\cite{wang2022internvideo, zhang2022actionformer, zeng2019graph} for better prediction. Therefore, we explore three solutions for ANet: (1) Directly regressing the temporal segments for 200 action categories; (2) with score fusion, averaging the normalized text embeddings of whole categories for action proposals; and (3) with score fusion, using the phrase, ``Someone is carrying out an activity'' for action proposals. By default, we adopt the second solution. A detailed comparison of these methods is provided in supplementary material.

\vspace{1mm}
\noindent\textbf{Training.}  
The training of three paired datasets utilizes a consistent AdamW optimizer and follows the same loss weights ($\lambda_{tad}=3$, $\lambda_{mr}=1$). Supplementary material will illustrate the slight differences in other settings.
% , including batch size, epochs, and learning rate.

\subsection{Ablation Study}
\label{subsec:ablation}

\begin{table}[ht]
\centering
\caption{Ablation studies on regression head, task-unified learning, loss weight, and data volumes. The best results are in \textbf{bold} and second best \ul{underlined}.}
\label{tab:ablation}
% \vspace{-3mm}
% 第一行子表
\begin{subtable}{.45\linewidth}
\centering
% \label{tab:ablation_reg_head}
\resizebox{1.0\linewidth}{!}{%
\begin{tabular}{cc|cc|cc}
                                  &                           & \multicolumn{2}{c|}{TAD} & \multicolumn{2}{c}{MR} \\ \hline
reg\_head                          & $\text{\#}FC_{reg}$                      & mAP         & R1@50      & R1@30      & R1@50     \\ \hline
\cellcolor[HTML]{EFEFEF}w/o query & -                         & 21.86       & 38.86      & 12.49      & 8.34      \\
                                  & \cellcolor[HTML]{EFEFEF}1 & 22.18       & 41.09      & 12.65      & 8.93      \\
                                  & \cellcolor[HTML]{EFEFEF}2 & 22.28       & 41.10      & 12.78      & 9.04      \\
\multirow{-3}{*}{w query} & \cellcolor[HTML]{EFEFEF}3 & \textbf{22.61} & \textbf{41.18} & \textbf{13.99} & \textbf{9.34}
\end{tabular}%
}
\caption{Effect of \textbf{query dependent regression head} evaluated on the Ego4D validation set under individual training. We investigate the effectiveness of queries in the regression head (reg\_head) and the number of fc layers in the MLP of the regression head ($\text{\#}FC_{reg}$).}
\label{tab:ablation_reg_head}
\end{subtable}
\hfill
\begin{subtable}{.50\linewidth}
\centering

\resizebox{0.85\linewidth}{!}{%
\begin{tabular}{cc|cc|cc}
\multirow{2}{*}{Method} & \multirow{2}{*}{policy} & \multicolumn{2}{c|}{TAD}        & \multicolumn{2}{c}{MR}                                            \\
       &          & mAP         & R1@50       & R1@30 & R1@50 \\ \hline
TAD    &         & 22.61       & 41.18       & -     & -         \\
MR     & \multirow{-2}{*}{individual}        & -           & -           & 13.99 & 9.34   \\ \hline
MR$\rightarrow$TAD &  & \cellcolor[HTML]{EFEFEF} 22.69    & \cellcolor[HTML]{EFEFEF} 41.59       & - & -   \\
TAD$\rightarrow$MR & \multirow{-2}{*}{pretrain} & -       & -       & 12.70 & 8.73   \\ \hline
 & Random   & \cellcolor[HTML]{EFEFEF} {\ul 22.78} &  \cellcolor[HTML]{EFEFEF} {\ul 43.08} & \cellcolor[HTML]{EFEFEF} 14.33 & \cellcolor[HTML]{EFEFEF} 9.73   \\
                  & Sync.                     & \cellcolor[HTML]{EFEFEF} \textbf{24.12} & \cellcolor[HTML]{EFEFEF} \textbf{44.14} &  \cellcolor[HTML]{EFEFEF} {\ul 14.48}    & \cellcolor[HTML]{EFEFEF} {\ul 10.09}        \\
\multirow{-3}{*}{TAD$+$MR}                  & Alt.                      & 21.73          & \cellcolor[HTML]{EFEFEF}41.32          & \cellcolor[HTML]{EFEFEF} \textbf{15.46} & \cellcolor[HTML]{EFEFEF} \textbf{10.56}  
\end{tabular}%
}
\caption{Effective of \textbf{task fusion learning} evaluated on the Ego4D validation set. For simplicity, synchronized task sampling and alternating task sampling are abbreviated as ``Sync.'' and ``Alt.''.}
\label{tab:ablation_cotrain}
% \vspace{-3mm}
\end{subtable}

% \vspace{-2mm} % 垂直间距

% 第二行子表
\begin{subtable}{.44\linewidth}
\centering
\resizebox{0.92\linewidth}{!}{%
\begin{tabular}{c|cc|cc}
loss weight & \multicolumn{2}{c|}{TAD}        & \multicolumn{2}{c}{MR}          \\
TAD-MR      & mAP            & R1@50         & R1@30          & R1@50          \\ \hline
1-3         & 19.61          & 40.01          & 14.12          & 9.55           \\
1-2         & 20.09          & 38.75          & 13.76          & 9.63           \\
1-1         & 22.55          & 40.50          & 14.33          & 9.86           \\
2-1         & {\ul{23.25}}          & {\ul{43.02}}          & {\ul{14.35}}          & \textbf{10.12} \\
3-1         & \textbf{24.12} & \textbf{44.14} & \textbf{14.48} & {\ul{10.09}}         
\end{tabular}%
}
\caption{Effect of \textbf{loss weights} on validation of Ego4D, using synchronized task sampling.}
\label{tab:ablation_loss_weight}
\end{subtable}
\hfill
\begin{subtable}{.55\linewidth}
\centering
\resizebox{0.84\linewidth}{!}{%

\begin{tabular}{cc|cc|cc}
\multicolumn{1}{c}{co-train} & \multicolumn{1}{c|}{data ratio} & \multicolumn{2}{c|}{TAD}        & \multicolumn{2}{c}{MR}                                \\
 policy      &    TAD-MR         & mAP         & mAP@50            & R5@50          & R5@70          \\ \hline
    & 1-0           & 38.60       & 58.31           & -              & -              \\
\multirow{-2}{*}{dedicated}     & 0-1           & -           & -                 & 77.28          & 53.86          \\ \hline
     & 0.25-0.25  & 36.78  & 56.96   & \cellcolor[HTML]{EFEFEF} 79.72  & \cellcolor[HTML]{EFEFEF} 54.43    \\  
     & 0.5-0.5  & \cellcolor[HTML]{EFEFEF} 38.62  & \cellcolor[HTML]{EFEFEF} 58.80   & \cellcolor[HTML]{EFEFEF} 80.12  & \cellcolor[HTML]{EFEFEF} 55.00    \\
     & 0.75-0.75  & \cellcolor[HTML]{EFEFEF} {\ul{39.24}}  & \cellcolor[HTML]{EFEFEF} {\ul{59.28}}   & \cellcolor[HTML]{EFEFEF} {\ul{80.53}}  & \cellcolor[HTML]{EFEFEF} {\ul{56.50}}    \\ 
\multirow{-4}{*}{Random} & 1-1      & \cellcolor[HTML]{EFEFEF} \textbf{39.82} &  \cellcolor[HTML]{EFEFEF} \textbf{60.04}           & \cellcolor[HTML]{EFEFEF} \textbf{80.83} & \cellcolor[HTML]{EFEFEF} \textbf{57.21} \\
\end{tabular}%
}
\caption{Impact of various ratios of \textbf{data volumes} evaluated on validation of ANet and val\_2 of ANet-Caption.}
\label{tab:ablation_data_ratio}

\end{subtable}
% \vspace{-8mm}
\end{table}

In this section, we conduct ablation experiments on the proposed query dependent head, the task fusion learning, loss weights of two tasks, and data volumes in the training set, as presented in Table~\ref{tab:ablation}.

\noindent\textbf{Query dependent regression head.} As shown in Table~\ref{tab:ablation_reg_head}, the regression head with queries as input yields better performance for both tasks (21.86\%$\rightarrow$22.61\% mAP in TAD and 12.49\%$\rightarrow$13.99\% R1@30 in MR). This enhancement is attributed to the semantic information of the query, which enables more precise time intervals for specific actions. In the query transformation branch of this head, the MLP that is equipped with 3 fc layers achieves the best performance.

\noindent\textbf{Task fusion learning.} The comparison of pre-training and co-training on the Ego4D validation set is presented in Table~\ref{tab:ablation_cotrain}. We employ the TAD task for pre-training and subsequent fine-tuning on the MR task, denoted as ``TAD$\rightarrow$MR''. The reverse direction is indicated as ``MR$\rightarrow$TAD''. The effect of pre-training varied for different tasks. For ``MR$\rightarrow$TAD'', there is a slight improvement of 0.08\% mAP in TAD. However, using TAD as the pre-training task negatively affects MR, possibly due to overfitting to the TAD domain. 
Furthermore, we analyse the efficacy of three sampling methods in co-training: Synchronized task sampling (``Sync.''), Alternating task sampling (``Alt.''), and Random task sampling (``Random''). It is worth noting that the co-trained models utilize \textbf{a shared set of parameters} for testing both tasks. Firstly, the ``Sync.'' method yields significant improvements in the co-training results for both TAD and MR tasks. Specifically, Ego4D sees an increase of 1.51\% mAP in TAD and 0.75\% R1@50 in MR. Next, the ``Alt.'' method leads to significant enhancements in MR (+1.22\% R1@50), surpassing the effects of the ``Sync.'' method. The reason for this is that separating TAD and MR into alternative iterations effectively treats TAD as an MR task, thereby increasing the number of annotations and negative samples for MR. However, in terms of TAD metrics, the ``Alt.'' biases the results towards MR, unlike the ``Sync.'' method which better utilizes the TAD and MR annotations to achieve mutually beneficial results. Regarding the ``Random'' method, it shows improvements in both TAD and MR, with the effects lying between the ``Sync.'' and ``Alt.''. More experiments on task fusion learning for Charades and ANet will be presented in the supplementary material.

\noindent\textbf{Loss weights.} The impact of different loss weights on the outcomes is significant, as indicated by Table~\ref{tab:ablation_loss_weight}. For example, when using a ``1-1'' loss weight, the co-training performance is inferior to the dedicated model for TAD (22.55 \vs. 22.61 mAP). Therefore, we carefully choose a loss weight of ``3-1'' to ensure that the co-train models are trained with a suitable balance of tasks.

\noindent\textbf{Data volumes.} The impact of data volumes on the validation of ANet is presented in Table~\ref{tab:ablation_data_ratio}, By controlling the number of training data for each task (\eg 0.5-0.5 means only half of the videos from TAD and MR), we observe a positive effect on both tasks with an increase in data volumes, as shown in Fig.~\ref{fig:intro_2}. \emph{Noteworthily}, in TAD, the co-trained model using only \textbf{50\% of the training videos} slightly outperforms the dedicated model, while in MR, even using only \textbf{25\% of the training videos} significantly surpasses the dedicated model.
It reveals that the mutual benefits are not merely derived from the increased quantity of annotations, but rather from the enhanced effectiveness of co-training.

\subsection{Comparison with the state-of-the-art}

\begin{table}[t]
\caption{Comparison with the state-of-the-art on Ego4D validation set and test set. 
}
\centering
% \vspace{-2.5mm}
\resizebox{11.0cm}{!}{%
\setlength\tabcolsep{4.pt}
\begin{tabular}{c|lc|cc|cccc} &   & &    \multicolumn{2}{c|}{TAD} & \multicolumn{4}{c}{MR}       \\
& \multirow{-2}{*}{Method} & \multirow{-2}{*}{Feature} &
  mAP & R1@50 & R1@30 & R1@50 & R5@30 & R5@50 \\ \hline
& EgoVLP (VSGN)\cite{lin2022egocentric}          &  EgoVLP        & 10.69       & 29.59      & -     & -    & -     & -     \\
& InternVideo (AF)\cite{chen2022ego4d}       & VideoMAE-verb & 20.69       & 37.24      & -     & -    & -     & -     \\ 
& ActionFormer\cite{zhang2022actionformer}       & Ensemble & {\ul{21.40}}       & {\ul{38.73}}      & -     & -    & -     & -     \\ 
& EgoVLP (VSLNet)\cite{lin2022egocentric}        &  EgoVLP        & -           & -          & 10.84 & 6.81 & 18.84 & 13.45 \\
& InternVideo (VSLNet)\cite{chen2022ego4d}   & VideoMAE-verb & -           & -          & 12.32 & 7.43 & 21.73 & 15.07 \\
& InternVideo (VSLNet)\cite{chen2022ego4d}   & VideoMAE-noun & -           & -          & {\ul{12.78}} & {\ul{8.08}} & {\ul{21.89}} & {\ul{15.64}} \\
\multirow{-7}{*}{val} & \cellcolor[HTML]{EFEFEF} UniMD+Sync. &
  \cellcolor[HTML]{EFEFEF} VideoMAE-verb &
  \cellcolor[HTML]{EFEFEF} \textbf{24.12} &
  \cellcolor[HTML]{EFEFEF} \textbf{44.14} &
  \cellcolor[HTML]{EFEFEF} \textbf{14.48} &
  \cellcolor[HTML]{EFEFEF} \textbf{10.09} &
  \cellcolor[HTML]{EFEFEF} \textbf{35.00} &
  \cellcolor[HTML]{EFEFEF} \textbf{25.04} \\ \hline
& EgoVLP (VSGN)\cite{lin2022egocentric}         &  EgoVLP        & 9.78        & 27.98      & -     & -    & -     & -     \\
& InternVideo (AF)\cite{chen2022ego4d}      & VideoMAE-verb & 19.31       & 35.58      & -     & -    & -     & -     \\
& ActionFormer\cite{zhang2022actionformer}       & Ensemble & {\ul{21.76}}       & {\ul{42.54}}     & -     & -    & -     & -     \\ 
& EgoVLP (VSLNet)\cite{lin2022egocentric}       &  EgoVLP        & -           & -          & 10.46 & 6.24 & 16.76 & 11.29 \\
& InternVideo (VSLNet)\cite{chen2022ego4d}  & VideoMAE-verb & -           & -          & {\ul{13.03}} & {\ul{7.87}} & {\ul{20.32}} & {\ul{13.29}} \\

% \rowcolor[HTML]{EFEFEF}
\multirow{-6}{*}{test} 
& \cellcolor[HTML]{EFEFEF} {UniMD+Sync.} &
  \cellcolor[HTML]{EFEFEF} VideoMAE-verb &
  \cellcolor[HTML]{EFEFEF} \textbf{23.25} &
  \cellcolor[HTML]{EFEFEF} \textbf{44.80} &
  \cellcolor[HTML]{EFEFEF} \textbf{14.16} &
  \cellcolor[HTML]{EFEFEF} \textbf{10.06} &
  \cellcolor[HTML]{EFEFEF} \textbf{26.95} &
  \cellcolor[HTML]{EFEFEF} \textbf{19.16} 
\end{tabular}%
}

\label{tab:ego4d_sota}
% \vspace{-2.0mm}

\end{table}

\begin{table}[t]
\caption{Comparison with state-of-the-art methods evaluated on test set of Charades for TAD and Charades-STA for MR. 
}
% \vspace{-3.5mm}
\centering
\resizebox{10.0cm}{!}{%
\setlength\tabcolsep{4.5pt}
\begin{tabular}{lc|c|cccc}
                    &               & TAD            & \multicolumn{4}{c}{MR}              \\
\multirow{-2}{*}{Method} &
  \multirow{-2}{*}{Feature} &
  mAP &
  R1@50 &
  R1@70 &
  R5@50 &
  R5@70 \\ \hline
TGM\cite{piergiovanni2019temporal}                 & I3D           & 22.3           & -     & -     & -           & -     \\
MLAD\cite{tirupattur2021modeling}                & I3D           & 22.9           & -     & -     & -           & -     \\
biGRU+VS-ST-MPNN\cite{mavroudi2020representation}    & I3D           & 23.7           & -     & -     & -           & -     \\
Coarse-Fine\cite{kahatapitiya2021coarse}         & -             & 25.10          & -     & -     & -           & -     \\
MS-TCT\cite{dai2022ms}              & I3D (RGB) & 25.4           & -     & -     & -           & -     \\
PDAN\cite{dai2021pdan}                & I3D           & 26.5           & -     & -     & -           & -     \\
CTRN\cite{dai2021ctrn}         & I3D           & \textbf{27.8}            & -     & -     & -           & -     \\ \hline
UMT (video, audio)\cite{liu2022umt}   & VGG           & -              & 48.31 & 29.25 & 88.79       & 56.08 \\
UMT (video, optical)\cite{liu2022umt} & VGG           & -              & 49.35 & 26.16 & 89.41       & 54.95 \\
MomentDETR\cite{lei2021detecting}          & SLOWFAST+CLIP & -              & 53.63 & 31.37 & -           & -     \\
MomentDETR (pt)\cite{lei2021detecting}      & SLOWFAST+CLIP & -              & 55.65 & 34.17 & -           & -     \\
QD-DETR\cite{moon2023query}             & SLOWFAST+CLIP & -              & 57.31 & 32.55 & -           & -     \\
UnLoc-L\cite{yan2023unloc}             & CLIP          & -              & 60.8  & 38.4  & 88.2        & 61.1  \\ \hline
\rowcolor[HTML]{EFEFEF} 
UniMD+Sync.   & CLIP           & 19.72 & 55.67 & 34.89 & 89.54 & 59.85 \\ 
\rowcolor[HTML]{EFEFEF} 
UniMD+Sync.   & I3D           & 24.06 & {\ul{63.90}} & {\ul{42.22}} & \textbf{92.12} & {\ul{67.23}} \\
\rowcolor[HTML]{EFEFEF} 
UniMD+Sync.   & I3D+CLIP      & {\ul{26.53}} & \textbf{63.98} & \textbf{44.46} & {\ul{91.94}}       & \textbf{67.72} 
\end{tabular}%
}
% \vspace{-3.0mm}

\label{tab:charades_sota}
\end{table}

\begin{table}[t]
\caption{Comparison with state-of-the-art on the validation set of ANet for TAD and the val\_2 set of ANet-Caption for MR. The features used in \textcolor{gray}{PRN$^{\dag}$} cannot be obtained.}
% \vspace{-3.5 mm}
\centering
\resizebox{8cm}{!}{%
\setlength\tabcolsep{4.5pt}
\begin{tabular}{lc|cc|cccc}
                       &             & \multicolumn{2}{c|}{TAD} & \multicolumn{2}{c}{MR}                 \\
\multirow{-2}{*}{Method} & \multirow{-2}{*}{Feature} & mAP            & mAP@50                 & R5@50          & R5@70          \\ \hline
TSP\cite{alwassel2021tsp}                    & TSP         & 35.81       & 51.26           & -     & -     \\
VSGN\cite{zhao2021video}                   & TSP         & 35.94       & 53.26       & -     & -     \\
ActionFormer\cite{zhang2022actionformer}           & TSP         & 36.6        & 54.7      & -     & -     \\
TadTR\cite{liu2022end}                  & TSP         & 36.75       & 53.62     & -     & -     \\
TriDet\cite{shi2023tridet}                 & TSP         & 36.8        & 54.7    & -     & -     \\
TCANet\cite{qing2021temporal}                 & SLOWFAST    & 37.56       & 54.33      & -     & -     \\
InternVideo (AF)\cite{wang2022internvideo}        & InternVideo & 39.0        & -           & -     & -     \\
\textcolor{gray}{PRN$^{\dag}$}\cite{wang2021proposal}                    & \textcolor{gray}{CSN}         & \textcolor{gray}{\ul{39.4}}        & \textcolor{gray}{57.9}         & -     & -     \\ 
UnLoc-L\cite{yan2023unloc} & CLIP        & -           & {\ul{59.3}}       & -            & -     \\ \hline
2D TAN\cite{zhang2020learning}                 & C3D         & -           & -      & 77.1     & {\ul{62.0}}     \\
DRN\cite{zeng2020dense}                    & C3D         & -           & -           & 77.97 & 50.30 \\
VLGNet\cite{soldan2021vlg}                   & C3D                       & -                    &  -  & 77.15          & \textbf{63.33} \\
UnLoc-L\cite{yan2023unloc}                  & CLIP                      & -              & -               & 79.2           & 61.3     \\ \hline
\rowcolor[HTML]{EFEFEF} 
UniMD+Sync.        & TSP               & 38.76 & 58.21          & {\ul{79.98}}    & 55.78          \\
\rowcolor[HTML]{EFEFEF} 
UniMD+Sync.        & InternVideo               & \textbf{39.83} & \textbf{60.29}    & \textbf{80.54}    & 57.04                
\end{tabular}%
}
\label{tab:anet_sota}
% \vspace{-4.0 mm}
\end{table}

This section compares the state-of-the-art (SOTA) methods for TAD and MR on three paired datasets. \emph{The purpose of this comparison is to examine the impact of mutual benefits based on various video features, rather than aiming for SOTA.}

\vspace{1mm}
\noindent\textbf{Results on Ego4D.} The comparison between our method and the SOTA approaches is presented in Table~\ref{tab:ego4d_sota}. Our models exhibit superior performance on both validation and test sets, establishing a new SOTA result for TAD and MR tasks. Specifically, when compared to InternVideo (AF)~\cite{chen2022ego4d} which uses the same video features, 
our method (referred to as ``UniMD+Sync.'') achieves an increase of 3.94\% mAP and 9.22\% R1@50 in the TAD test set. Despite using inferior features compared to ActionFormer\cite{zhang2022actionformer}, our method outperforms the SOTA by 1.49\% mAP and 2.26\% R1@50. 
Additionally, in the MR test set, UniMD surpasses InternVideo (VSLNet)~\cite{chen2022ego4d} with a 1.13\% improvement in R1@30 and a remarkable 6.63\% increase in R5@30. Last but not least, UniMD is capable of performing both TAD and MR tasks \textbf{in a single model}, with a negligible increase in computational cost, \eg, less than 1\% compared to ActionFormer.

\vspace{1mm}
\noindent\textbf{Results on Charades and Charades-STA.} 
Table~\ref{tab:charades_sota} shows the performance of TAD and MR. 
In the TAD benchmark, our model achieves 24.06\% mAP when utilizing the I3D feature, which is lower compared to CTRN~\cite{dai2021ctrn}. It is worth noting that CTRN is specifically designed for densely labelled action datasets, using a graph-based classifier to tackle the co-occurring action challenge. When incorporating additional CLIP features as video features, our approach shows significant improvement and achieves competitive performance with CTRN. In the MR benchmark, our model incorporating I3D and CLIP features achieves a new SOTA result of 63.98\% in R1@50, exhibiting a notable increase of 3.18\%.

\vspace{1mm}
% \subsubsection{Results on ActivityNet and ActivityNet-Caption}
\noindent\textbf{Results on ANet and ANet-Caption.} The comparisons in both TAD and MR tasks are shown in Table~\ref{tab:anet_sota}. Our co-trained model ranks at the top in the TAD benchmark with better performance than InternVideo (AF), when utilizing the same feature InternVideo.  
It sets a new SOTA result with 39.83\% mAP and 60.29 \% mAP@50 in the TAD task.
In the MR task, our method exhibits a superior performance of 80.54\% R5@50. 

\section{Conclusion}

This paper provides an answer to the question of ``whether the TAD and MR tasks can benefit each other by fusing them into a single model?''. To achieve this, we begin by designing a task-unified network, termed UniMD, with uniform interfaces for task input and output. We further explore various task fusion learning methods to enhance the collaboration between TAD and MR. Through experiments on three paired datasets, we present evidence that the task fusion learning approach effectively improves the performance of both tasks. 

% \clearpage\mbox{}Page \thepage\ of the manuscript.
% \clearpage\mbox{}Page \thepage\ of the manuscript.
% \clearpage\mbox{}Page \thepage\ of the manuscript.
% \clearpage\mbox{}Page \thepage\ of the manuscript.
% \clearpage\mbox{}Page \thepage\ of the manuscript. This is the last page.
% \par\vfill\par
% Now we have reached the maximum length of an ECCV \ECCVyear{} submission (excluding references).
% References should start immediately after the main text, but can continue past p.\ 14 if needed.
% \clearpage  % TODO REVIEW/FINAL: This \clearpage needs to be removed from both review and camera-ready versions.

% ---- Bibliography ----
%
% BibTeX users should specify bibliography style 'splncs04'.
% References will then be sorted and formatted in the correct style.
%
\bibliographystyle{splncs04}
\bibliography{main}



\section*{Supplementary material}

# UniMD: Towards Unifying Moment Retrieval and Temporal Action Detection

## Supplementary Material

| | | Temporal Action Detection | | | | | | Moment Retrieval | | | | | |
|---|---|---|---|---|---|---|---|---|---|---|---|---|---|
| | #clip | #clip$_{diff}$ | #cls | Avg$_{dura}$(s) | Avg$_{ins}$ | Avg$_{ins\_dura}$(s) | #clip$_{overlap}$ | #clip | #clip$_{diff}$ | #query | Avg$_{dura}$(s) | Avg$_{ins}$ | Avg$_{ins\_dura}$(s) |
| | | | | | | Ego4D | | | | | | | |
| train | 1.5k | 1.5k | | | | | 17 | 1k | 1k | 11.3k | | | |
| val | 0.5k | 0.5k | | | | | 7 | 0.3k | 0.3k | 3.9k | | | |
| total | 2k | 2k | 110 | 472 | 9.2 | 42.8 | **24** | 1.3k | 1.3k | 15.2k | 494 | 10.2 | 10.6 |
| | | | | | | Charades | | | | | | | |
| train | 8k | 2.6k | | | | | 5.3k | 5.3k | 0 | 12.4k | | | |
| val | 1.8k | 0.5k | | | | | 1.3k | 1.3k | 0 | 3.7k | | | |
| total | 9.8k | 3.1k | 157 | 30 | 6.8 | 12.9 | **6.6k** | 6.6k | 0 | 16.1k | 30 | 2.4 | 8.3 |
| | | | | | | ActivityNet | | | | | | | |
| train | 1w | | 15 | | | | 9972 | 1w | 37 | 3.7w | | | |
| val_1 | | | 9 | | | | 4917 | 0.5w | 0 | 1.7w | | | |
| val_2 | 0.5w | | 41 | | | | 4885 | 0.5w | 0 | 1.7w | | | |
| total | 2w | | 65 | 200 | 118 | 1.5 | 49 | **19774** | 2w | 37 | 7.1w | 118 | 3.5 | 35.5 |

Table A. **Statistics of three paired datasets that cover both TAD and MR tasks.** #clip denotes the number of video clips, #clip$_{diff}$ presents the number of non-overlapping videos, and #clip$_{overlap}$ refers to the number of overlapping videos. Avg$_{dura}$, Avg$_{ins\_dura}$, and Avg$_{ins}$ indicate the average video duration, the average instance duration, and the average number of instances per video, respectively. #query represents the number of queries in MR while #cls denotes the number of action categories in TAD. The number underlined in ActivityNet indicates that the validation set in TAD only has one version, while ActivityNet-Caption has two versions, val_1 and val_2.

## 8. Appendix

In the appendix, we describe (1) more details about three paired datasets (Section A), (2) additional ablation experiments (Section B), and (3) qualitative results (Section C). For sections, figures, tables, and equations, we use numbers (*e.g.* Sec. 1) to refer to the main paper and capital letters (*e.g.* Sec. A) to refer to this appendix.

## A. Details About Datasets

**Statistical analysis of datasets.** (1) **Ego4D** [16] is a large-scale first-person video dataset that consists of ∼3000 hours of egocentric videos. The TAD task of Ego4D includes 110 action categories. Specifically, the TAD task has an average of 9.2 instances per video across 2k videos, while the MR task has an average of 10.2 instances per video across 1.3k videos. (2) **Charades and Charades-STA.** Charades [46] is a large-scale dataset that consists of 9.8k videos, with an average duration of 30 seconds. The dataset includes a total of 157 action categories that are densely labeled in the videos. Charades-STA [15], an extension of Charades, introduces natural language descriptions as annotations. Charades-STA consists of 6.6k videos, with the same average duration as Charades. (3) **ANet and ANet-Caption.** ActivityNet (ANet) [3] contains 200 action categories and ∼20,000 videos. Additionally, ActivityNet-Caption (ANet-Caption) [25] is built upon ANet, including descriptions of events and their corresponding timestamps. ANet-Caption also comprises ∼20,000 videos, having the same average duration of 118 seconds as ANet.

However, **the proportion of overlapping videos between the TAD and MR tasks varies across datasets**, as shown in Table A. In Ego4D, there is a minimal overlap of only ∼2%. Conversely, ANet exhibits a substantial overlap, with more than 99% of the videos being shared. In the Charades, all of the videos from Charades-STA are included, while Charades-STA covers 67% of the videos in Charades. Since each pair of datasets shares the same video source, the duration of the videos in TAD is approximately identical to that in MR. In addition, there are notable differences in the number of instances and average instance duration. In Ego4D, despite having similar average instance numbers, there is approximately a 300% difference in average instance duration. In Charades, TAD has an average instance number 2.8 times that of MR, and its average instance duration is 50% longer than that of MR. ANet presents a different situation, where MR has a higher average instance number compared to TAD, but the average instance duration is shorter in MR.

**Annotation visualizations.** Figure A displays six additional annotation visualizations that illustrate the relationship between TAD and MR.

1

| Component | | | TAD | |
|---|---|---|---|---|
| Encoder | #$D_{reg}$ | $scale_{cls}$ | mAP | R1@50 |
| ConvNext | 128 | Y | 21.86 | 38.79 |
| | 256 | Y | 22.19 | 40.62 |
| | 512 | Y | **22.61** | **41.18** |
| | 512 | N | 22.05 | 38.58 |
| Transformer | 512 | Y | 21.40 | 38.90 |

Table B. Effect of **different network components** evaluated on the TAD task of Ego4D validation set under individual training. The TAD task is the Moment Query task in Ego4D. The #$D_{reg}$ means the dimensions in the query transformation branch of the regression head and the $scale_{cls}$ indicates the scale operation in the classification head.

| Method | solution | TAD | | MR | |
|---|---|---|---|---|---|
| | | mAP | mAP@50 | R5@50 | R5@70 |
| MR | - | - | - | 77.28 | 53.86 |
| TAD | 1 | 37.78 | 57.48 | - | - |
| TAD+MR | | 37.98 | 57.48 | 79.90 | 53.71 |
| TAD | 2† | 38.60 | 58.31 | - | - |
| TAD+MR | | **39.82** | **60.04** | **80.83** | **57.21** |
| TAD | 3† | 38.16 | 57.42 | - | - |
| TAD+MR | | 39.79 | 59.83 | 80.03 | 56.89 |

Table C. Effect of **the three solutions for TAD** evaluated on the validation set in ANet and val_2 split in ANet-Caption. Here, the methods referred to as "TAD+MR" utilize task fusion learning with random task sampling. The methods "TAD" and "MR" are dedicated models for each task. Solutions 2† and 3† utilize score fusion strategy.

## B. Additional Ablation Experiments

In this section, we explore the model settings of the vision encoder and decoder based on TAD metrics of Ego4D, as presented in Table B. Additionally, we investigate three solutions for ANet (Table C) and analyse the effect of task fusion learning on Charades (Table D) and ANet (Table E).

**The choice of encoder and decoder.** For the ablation study on the vision encoder, we conduct an experiment by replacing the ConvNext [35] blocks with Transformer units, following the same pipeline. As shown in Table B, using ConvNext as the main block of the vision encoder yields better performance (22.61% vs 21.40% in mAP) compared to Transformer. Additionally, we conduct an exploration of the query transformation branch in the regression head and the learnable scaling operation in the classification head. The comparison indicates that using a Multi-Layer Perceptron (MLP) with a channel size of 512 leads to superior semantic translation, resulting in a higher mAP. Furthermore, the integration of learnable scaling operations positively affects performance.

**Three solutions for ANet.** In the context of score fusion strategy in ANet, we explore three solutions for TAD, as outlined in Sec. 6.2. Based on the comparison presented in Table C, the score fusion strategy, implemented in solutions 2 and 3, enhances the performance of TAD compared to solution 1 which does not use external classification scores.

| Method | policy | TAD | MR | | | |
|---|---|---|---|---|---|---|
| | | mAP | R1@50 | R1@70 | R5@50 | R5@70 |
| TAD | dedicated | 22.31 | - | - | - | - |
| MR | | - | 60.19 | 41.02 | 91.61 | 65.86 |
| MR→TAD | pretrain | 22.61 | - | - | - | - |
| TAD→MR | | - | 62.66 | 43.15 | 90.86 | 64.49 |
| TAD+MR | Random | 23.82 | 63.41 | 42.42 | 92.34 | 67.74 |
| TAD+MR | Sync. | **24.06** | 63.90 | 42.22 | 92.12 | 67.23 |
| TAD+MR | Alt. | 21.36 | **64.89** | **43.76** | 92.02 | 66.48 |

Table D. Effect of **task fusion learning** evaluated on Charades and Charades-STA test set.

| Method | policy | TAD | | MR | |
|---|---|---|---|---|---|
| | | mAP | mAP@50 | R5@50 | R5@70 |
| TAD | dedicated | 38.60 | 58.31 | - | - |
| MR | | - | - | 77.28 | 53.86 |
| MR→TAD | pretrain | 39.11 | 59.46 | - | - |
| TAD→MR | | - | - | 80.68 | 55.98 |
| TAD+MR | Random | 39.82 | 60.04 | **80.83** | **57.21** |
| TAD+MR | Sync. | **39.83** | **60.29** | 80.54 | 57.04 |
| TAD+MR | Alt. | 39.66 | 59.51 | **80.83** | 57.08 |

Table E. Effect of **task fusion learning** evaluated on validation set of ANet for TAD and val_2 set of ANet-Caption for MR.

Besides, among the results of task fusion learning, this approach offers an advantage. Specifically, solution 2 which averages the text embeddings of entire categories, shows slightly better performance (mAP and mAP@50 in TAD, and R5@50 in MR) compared to the manual prompt for action proposals in solution 3.

**Task fusion learning on Charades and ANet.** We examine the impact of task fusion learning and different sampling methods on Charades (Table D) and ANet (Table E). In terms of pre-training, both Charades and ANet benefit from this approach, except for the R5@50 and R5@70 in the MR task of Charades. Regarding the effect of co-training on these two datasets, our results align with Ego4D's findings in Sec. 6.3. The synchronized task sampling method (referred to as "Sync.") yields significant improvements in the co-training results for both TAD and MR tasks. Specifically, Charades shows an increase of +1.75% mAP in TAD and +3.71% R1@50 in MR, whole ANet sees an increase of +1.23% mAP in TAD and +3.26% R5@50 in MR. Additionally, we explore the effect of the alternating task sampling (referred to as "Alt."). This sampling leads to significant enhancements in MR for both datasets (+4.70% R1@50 of Charades, +3.55% R5@50 of ANet).

## C. Qualitative Results

This section showcases the qualitative results obtained from videos that encompass both TAD and MR tasks, as depicted in Figure B. In our work, we convert the predefined action categories of TAD into natural language queries and treat these queries the same way as queries from MR. The qualitative results demonstrate the capability of unified moment detection using natural language descriptions from both TAD and MR.

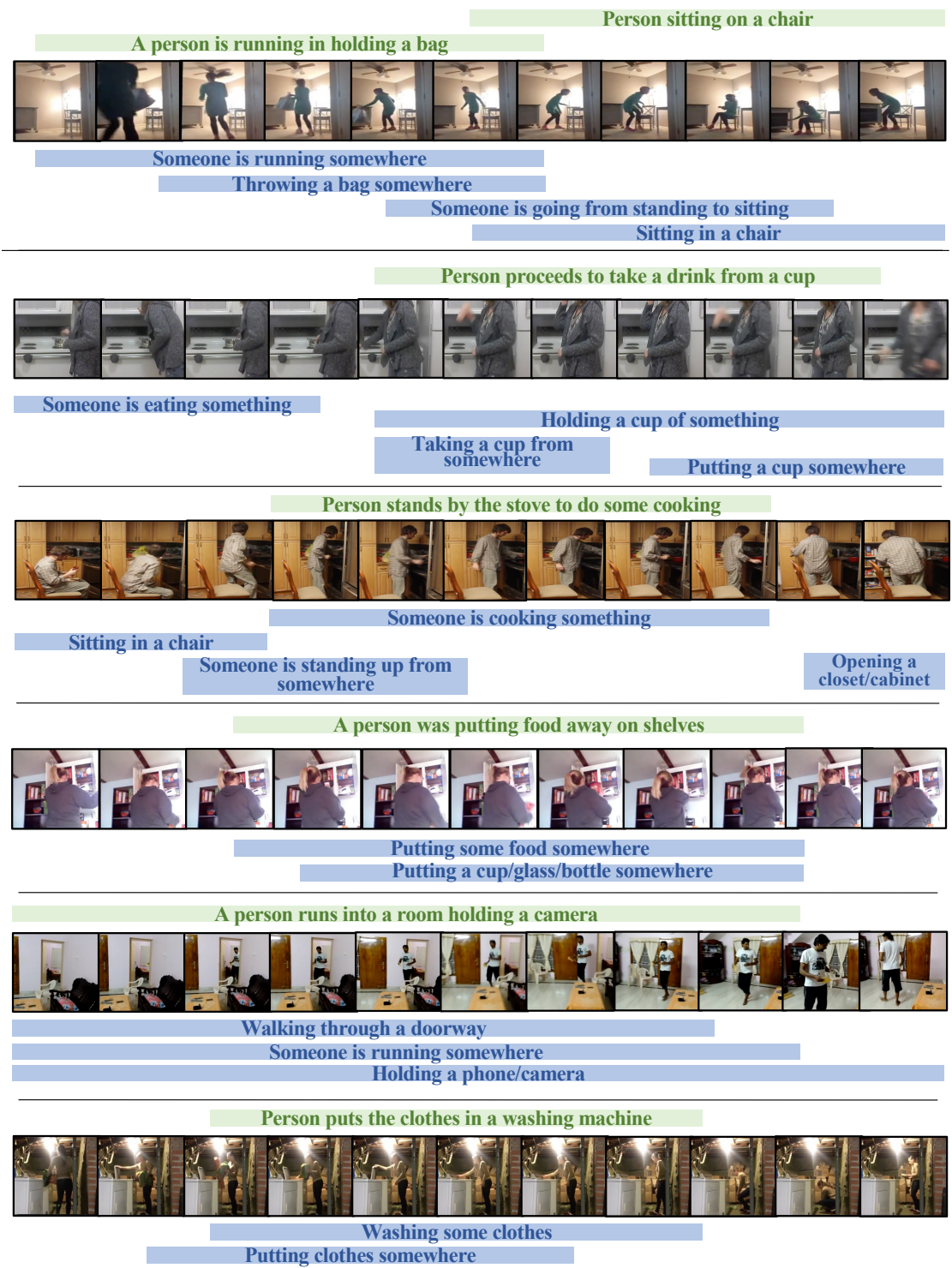

Figure A. **Visualization of annotations from Charades and Charades-STA.** The events in green presented above the videos are natural language descriptions from MR, while the actions in blue displayed below the videos belong to predefined categories of TAD.

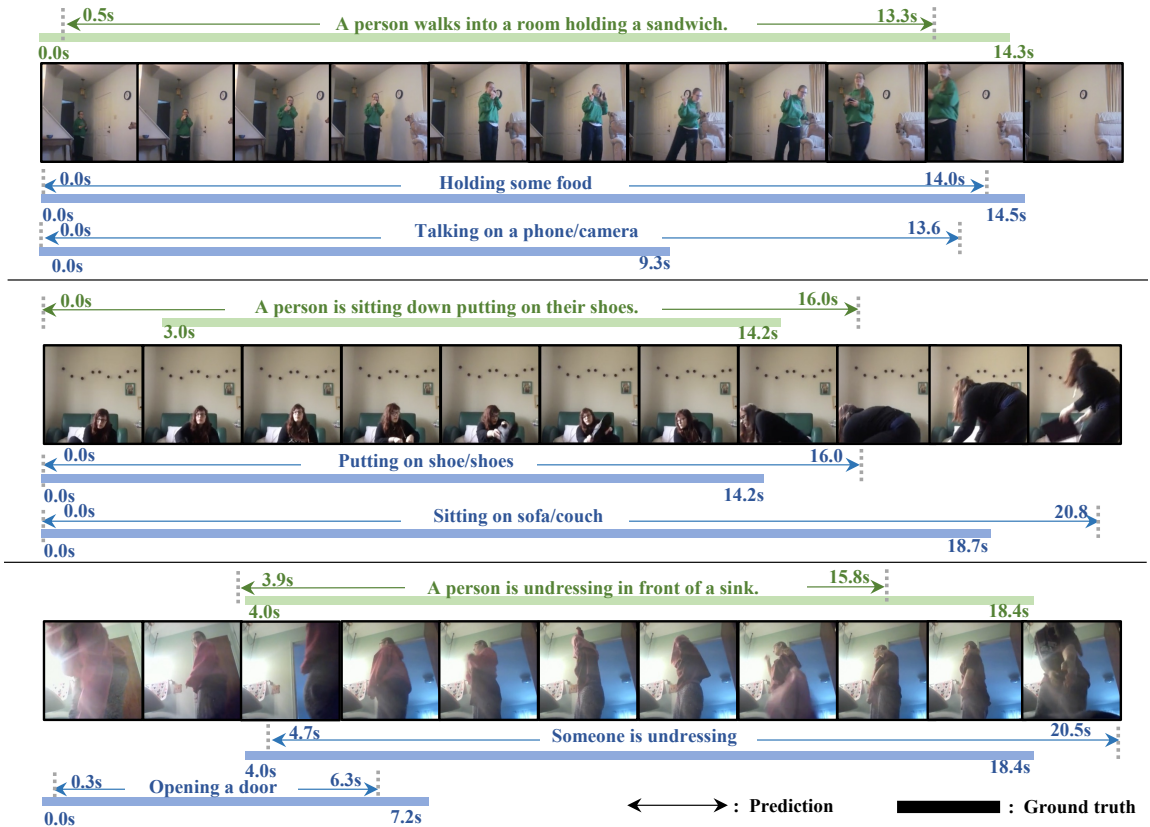

(a) Qualitative results on Charades and Charades-STA.

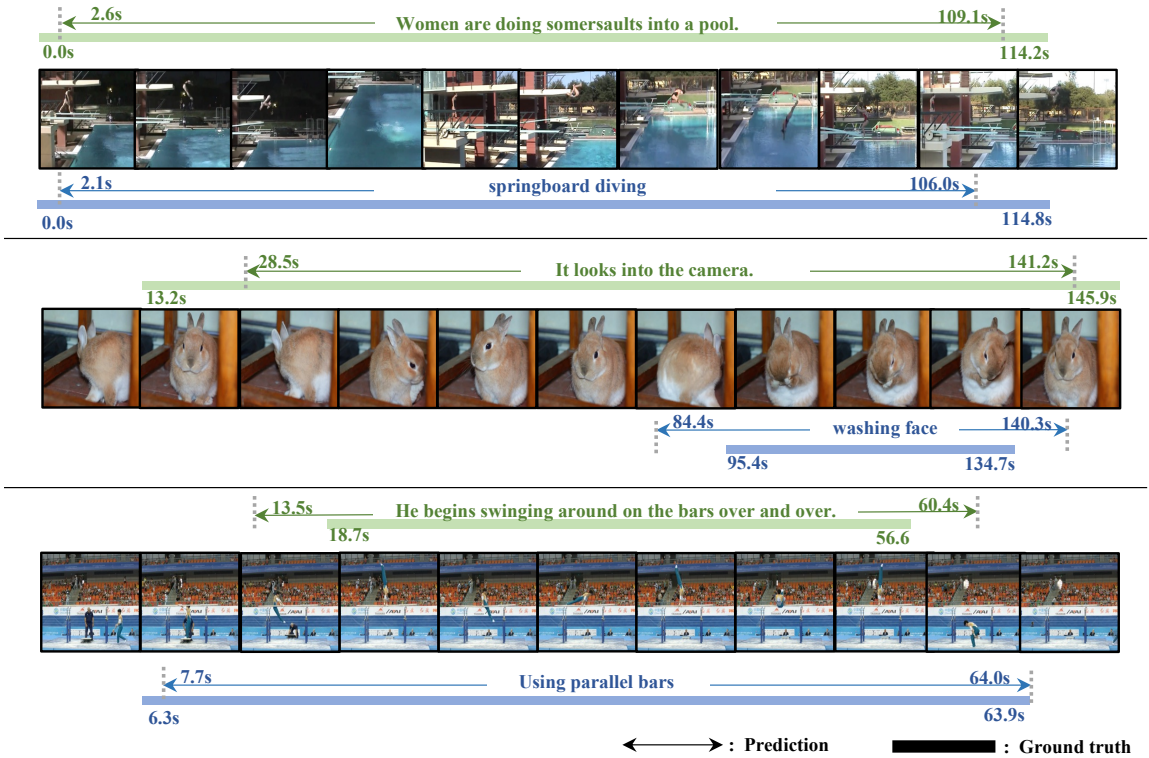

(b) Qualitative results on ANet and ANet-Caption.

Figure B. **Qualitative results**. The qualitative results provide evidence of the effectiveness of accurate boundary regression and moment detection using natural language descriptions from both MR and TAD. The queries in green presented above the videos are natural language descriptions from MR, while the queries in blue displayed below the videos belong to predefined categories of TAD.

4



\end{document}